\documentclass{article}

\usepackage[accepted]{icml2014}
\usepackage{amsfonts}
\usepackage{amsmath,amsthm}
\usepackage{amssymb}
\usepackage{algorithm}
\usepackage{graphicx}
\usepackage[noend]{algpseudocode}
\usepackage[caption=false]{subfig}

\newtheorem{theorem}{Theorem}[section]
\newtheorem{corollary}[theorem]{Corollary}

\newtheorem{lemma}[theorem]{Lemma}

\theoremstyle{definition}
\theoremstyle{definition}
\theoremstyle{observation}

\def\reals{\mathbb{R}}
\def\R{\reals}

\def\transp{^\textrm{T}}

\newcommand{\abs}[1]{\left|#1\right|}
\newcommand{\norm}[1]{\left\|#1\right\|}

\newcommand{\prob}[1]{{\sf Pr}\left(#1\right)}
\newcommand{\tr}[1]{\mathrm{tr} \left(#1\right)}

\newcommand{\E}[1]{\mathbb{E}\left(#1\right)}

\newcommand{\ip}[2]{\left<{ #1},{#2}\right>}

\newcommand{\TNorm}[1]{\ensuremath{\left\|#1\right\|_2}}

\renewcommand{\algorithmicrequire}{\textbf{Input:}}

\algrenewcommand{\algorithmicforall}{\textbf{for each}}
\newcommand{\ForEach}{\ForAll}

\graphicspath{{./figures/}}

\newcommand{\taba}{
\small{
\begin{tabular}{| c|| c| c| c| c| c| c|}
    \hline
    $\textrm{\textbf{a-}MNIST 1}$ & $300$ & $400$ & $500$ & $600$ & $700$\\
    \hline
    \hline
    RFM & $14.0$ & $12.3$ & $11.4$ & $10.3$ & $9.5$\\\hline
    TS & $13.1$ & $11.2$ & $10.0$ & $8.6$ & $8.0$\\\hline
    CM RFM & $\textbf{9.5}$ & $\textbf{7.7}$ & $\textbf{7.2}$ & $\textbf{6.6}$ & $\textbf{5.9}$\\\hline
    CM TS & $12.6$ & $10.8$ & $8.9$ & $7.9$ & $7.3$\\
    \hline
  \end{tabular}
  }
}

\newcommand{\tabb}{
\small{
\begin{tabular}{| c|| c| c| c| c| c| c|}
    \hline
    $\textrm{\textbf{b-}MNIST 2}$ & $2e12$ & $2e13$ & $2e14$ & $2e15$ & $2e16$\\
    \hline
    \hline
    RFM & $3.17$ & $2.30$ & $1.91$ & $1.62$ & $1.49$\\\hline
    TS & $3.25$ & $2.41$ & $2.01$ & $1.65$ & $1.41$\\\hline
    CM RFM & $3.09$ & $\textbf{2.18}$ & $1.79$ & $1.52$ & $1.27$\\\hline
    CM TS & $\textbf{2.90}$ & $2.20$ & $\textbf{1.75}$ & $\textbf{1.44}$ & $\textbf{1.12}$\\
    \hline
  \end{tabular}
  }
}

\newcommand{\tabc}{
\small{
\begin{tabular}{| c|| c| c| c| c| c| c|}
    \hline
    $\textrm{\textbf{c-}USPS}$ & $2e10$ & $2e11$ & $2e12$ & $2e13$ & $2e14$\\
    \hline
    \hline
    RFM & $5.97$ & $5.33$ & $4.68$ & $4.48$ & $4.13$\\\hline
    TS & $5.92$ & $5.03$ & $4.63$ & $4.48$ & $4.33$\\\hline
    CM RFM & $\textbf{5.68}$ & $\textbf{5.03}$ & $4.48$ & $4.28$ & $4.03$\\\hline
    CM TS & $5.77$ & $\textbf{5.03}$ & $\textbf{4.28}$ & $\textbf{4.23}$ & $\textbf{3.93}$\\
    \hline
  \end{tabular}
  }
}

\newcommand{\tabd}{
\small{
\begin{tabular}{| c|| c| c| c| c| c| c|}
    \hline
    $\textrm{\textbf{d-}COIL100}$ & $2e11$ & $2e12$ & $2e13$ & $2e14$ & $2e15$\\
    \hline
    \hline
    RFM & $11.11$ & $7.55$ & $6.33$ & $5.05$ & $4.83$\\\hline
    TS & $10.08$ & $7.19$ & $5.69$ & $4.75$ & $4.27$\\\hline
    CM RFM & $8.94$ & $6.86$ & $5.47$ & $4.52$ & $4.08$\\\hline
    CM TS & $\textbf{8.16}$ & $\textbf{5.97}$ & $\textbf{4.75}$ & $\textbf{4.02}$ & $\textbf{3.96}$\\
    \hline
  \end{tabular}
  }
}

\newcommand{\tabe}{
\small{
\begin{tabular}{| c|| c| c| c| c| c| c|}
    \hline
    $\textrm{\textbf{e-}PENDIGITS}$ & $2e6$ & $2e7$ & $2e8$ & $2e9$ & $2e10$\\
    \hline
    \hline
    RFM & $7.94$ & $3.94$ & $2.85$ & $2.28$ & $1.91$\\\hline
    TS & $11.20$ & $4.57$ & $2.37$ & $1.80$ & $1.77$\\\hline
    CM RFM & $\textbf{7.43}$ & $\textbf{3.57}$ & $\textbf{2.28}$ & $\textbf{1.97}$ & $\textbf{1.57}$\\\hline
    CM TS & $8.03$ & $3.80$ & $2.37$ & $2.05$ & $1.74$\\
    \hline
  \end{tabular}
  }
}

\icmltitlerunning{Compact Random Feature Maps}

\begin{document}

\twocolumn[
\icmltitle{Compact Random Feature Maps}

\icmlauthor{Raffay Hamid}{raffay@gmail.com}
\icmladdress{eBay Research Laboratory}
\icmlauthor{Ying Xiao}{ying.xiao@gatech.edu}
\icmladdress{Georgia Institute of Technology}
\icmlauthor{Alex Gittens}{agittens@ebay.com}
\icmladdress{eBay Research Laboratory}
\icmlauthor{Dennis DeCoste}{ddecoste@ebay.com}
\icmladdress{eBay Research Laboratory}

\icmlkeywords{kernel methods, kernel approximation, kernel regression, random features}

\vskip 0.3in
]

\begin{abstract}

Kernel approximation using randomized feature maps has recently gained a lot of interest. In this work, we identify that previous approaches for
polynomial kernel approximation create maps that are rank deficient, and therefore do not utilize the capacity of the projected feature space effectively.
To address this challenge, we propose compact random feature maps (CRAFTMaps) to approximate polynomial kernels more concisely and accurately.
We prove the error bounds of CRAFTMaps demonstrating their superior kernel reconstruction performance compared to the
previous approximation schemes. We show how structured random matrices can be used to efficiently generate CRAFTMaps, and present a single-pass
algorithm using CRAFTMaps to learn non-linear multi-class classifiers. We present experiments on multiple standard data-sets with performance
competitive with state-of-the-art results.

\end{abstract} 
\section{Introduction}

Kernel methods allow implicitly learning non-linear functions using explicit linear feature spaces~\cite{scholkopf1999advances}. These explicit feature spaces are typically high dimensional, and often pose what is called the \textit{curse of dimensionality}. One solution to this problem is the well known \textit{kernel trick}~\cite{aizerman1964theoretical}, where instead of directly learning a hyperplane classifier in $\mathbb{R}^d$, one considers a non-linear mapping $\Phi: \mathbb{R}^d \rightarrow \mathcal{H}$, such that for all $\mathbf{x}, \mathbf{y} \in \mathbb{R}^d, \langle \Phi(\mathbf{x}),\Phi(\mathbf{y}) \rangle_\mathcal{H} = \textrm{K}(\mathbf{x},\mathbf{y})$ for some kernel $\textrm{K}(\mathbf{x},\mathbf{y})$. One then learns a classifier $\mathbf{H}: \mathbf{x} \mapsto \mathbf{w}^\textrm{T} \Phi(\mathbf{x})$ for some $\mathbf{w} \in \mathcal{H}$.

 It has been observed however that with increase in training data size, the support of the vector $\mathbf{w}$ can undergo unbounded growth, which can result in increased training as well as testing time~\cite{steinwart2003sparseness}~\cite{bengio2006curse}. Previous approaches to address this \textit{curse of support} have mostly focused on embedding the non-linear feature space $\mathcal{H}$ into a low dimensional Euclidean space while incurring an arbitrarily small distortion in the inner product values~\cite{rahimi2007random}~\cite{kar2012random}~\cite{pham2013fast}. One way to do this is to construct a randomized feature map $\textbf{Z}: \mathbb{R}^d \rightarrow \mathbb{R}^\textrm{\textrm{D}}$ such that for all $\mathbf{x}, \mathbf{y} \in \mathbb{R}^d, \langle \textbf{Z}(\mathbf{x}), \textbf{Z}(\mathbf{y}) \rangle = \textrm{K}(\mathbf{x},\mathbf{y})$. Each component of $\textbf{Z}(\mathbf{x})$ can be computed by first projecting $\mathbf{x}$ onto a set of randomly generated $d$ dimensional vectors sampled from a zero-mean distribution, followed by computing the dot-products of the projections. While randomized feature maps are applicable to approximate the more general class of dot-product kernels, in this work we focus on analyzing polynomial kernels, where $\textrm{K}(\mathbf{x},\mathbf{y})$ is of the form $(\langle \mathbf{x}, \mathbf{y} \rangle$$+q)^r$, with $q \in \mathbb{N}_{0}$ and $r \in \mathbb{R}^{+}$.

It has been shown that $|\langle \textbf{Z}(\mathbf{x}), \textbf{Z}(\mathbf{y}) \rangle - \textrm{K}(\mathbf{x},\mathbf{y})|$ reduces exponentially as a function of $\textrm{D}$~\cite{kar2012random}~\cite{pham2013fast}. However in practice, to approximate $\textrm{K}(\mathbf{x},\mathbf{y})$ well, $\textrm{D}$ can still need to be increased to values that may not be amenable from the perspective of learning a classifier in $\mathbb{R}^{\textrm{D}}$. This is especially true for higher values of $r$. Furthermore, we show that the feature spaces constructed by random feature maps are over complete and rank deficient. This rank deficiency can in turn result in the under-utilization of the projected feature space from a learning perspective where the model parameters learned in $\mathbb{R}^{\textrm{D}}$ can have a significant number of components very close to zero.

This presents us with the dilemma whether to create feature maps that approximate exact kernel values accurately, or ones that enable efficient classifier learning. To resolve this dilemma, we propose compact random feature maps (CRAFTMaps) as a more concise representation of random feature maps that can approximate polynomial kernels more accurately. We show that the information content of $\textbf{Z}: \mathbb{R}^d \rightarrow \mathbb{R}^\textrm{D}$ can be captured more compactly by generating an alternate random feature map $\textbf{Q}: \mathbb{R}^\textrm{D} \rightarrow \mathbb{R}^{\textrm{E}}$, such that $\textrm{E} < \textrm{D}$, and $\langle \textbf{Q}(\textbf{Z}(\mathbf{x}))$, $\textbf{Q}(\textbf{Z}(\mathbf{y})) \rangle$ approximates $\langle \textbf{Z}(\mathbf{x}), \textbf{Z}(\mathbf{y}) \rangle$. CRAFTMaps are therefore constructed by first up projecting the original data non-linearly to $\mathbb{R}^{\textrm{D}}$ in order to minimize $|\langle \textbf{Z}(\mathbf{x}), \textbf{Z}(\mathbf{y}) \rangle - \textrm{K}(\mathbf{x},\mathbf{y})|$. This is followed by linearly down projecting the up-projected vectors to $\mathbb{R}^{\textrm{E}}$ with $\textrm{E} < \textrm{D}$ in order to capture the underlying structure in $\mathbb{R}^{\textrm{D}}$ more compactly. We present both analytical as well as empirical evidence of the fact that the ``up/down" projections employed by CRAFTMaps approximate $\textrm{K}(\mathbf{x},\mathbf{y})$ better than a direct random polynomial feature map $\textbf{Z}: \mathbb{R}^d \rightarrow \mathbb{R}^\textrm{E}$.

The additional cost of down projecting from $\mathbb{R}^{\textrm{D}}$ to $\mathbb{R}^{\textrm{E}}$ incurred by CRAFTMaps is well-justified by the efficiency gains they offer in terms of training in $\mathbb{R}^{\textrm{E}}$. To further improve the efficiency of CRAFTMaps, we show how they can be generated using structured random matrices, in particular Hadamard transform, that reduces the cost of multiplying two $n \times n$ matrices from $\mathcal{O}(n^{3})$ to $\mathcal{O}(n^{2}\rm{log}$$(n))$. This gain is exploited for both up as well as down projection steps of CRAFTMaps. Note that while down-projection using structured random matrices is straight forward~\cite{tropp2011improved}, we need to incorporate a few novel modifications to previous structured random projection approaches before they can be used for the up-projection step (see $\S$~\ref{ss:e_craftmaps}).

The compactness of CRAFTMaps makes them particularly suitable for using Hessian based methods to learn classifiers in a single pass over the data. Moreover, we show how CRAFTMaps can be used to learn multi-class classifiers in a streaming manner, using the previously proposed framework of error correcting output codes (ECOCs)~\cite{dietterich1995solving}, to minimize the least square error between the predicted and the true class labels. This combination of CRAFTMaps and ECOCs is particularly powerful as it can be formalized as a matrix-matrix multiplication, and can therefore maximally exploit the multi-core processing power of modern hardware using BLAS$3$~\cite{golub2012matrix}. Finally, by requiring minimal communication among \textit{mappers}, this framework is well-suited for \textit{map-reduce} based settings.

\section{Related Work}

Extending the kernel machines framework to large scale learning has been explored in a variety of ways~\cite{bottou2007large}. The most popular of these approaches are decomposition methods for solving Support Vector Machines~\cite{platt1999using}~\cite{chang2011libsvm}. While in general extremely useful, these methods do not always scale well to problems with more than a few hundreds of thousand data-points.

To solve this challenge, several schemes have been proposed to explicitly approximate the kernel matrix, including low-rank approximations~\cite{blum2006random}~\cite{bach2005predictive}, sampling individual entries~\cite{scholkopf2002sampling}, or discarding entire rows~\cite{drineas2005nystrom}. Similarly, fast nearest neighbor look-up methods have been used to approximate multiplication operations with the kernel matrix~\cite{shen2006fast}. Moreover, formulations leveraging concepts from computational geometry have been explored to obtain efficient approximate solutions for SVM learning~\cite{tsang2006core}.

An altogether different approximation approach that has recently gained much interest is to approximate the kernel function directly as opposed to explicitly operating on the kernel matrix. This can be done by embedding the non-linear kernel space into a low dimensional Euclidean space while incurring an arbitrarily small additive distortion in the inner product values~\cite{rahimi2007random}. By relying only on the embedded space dimensionality, this approach presents a potential solution to the aforementioned \textit{curse of support}, and is similar in spirit to previous efforts to avoid the \textit{curse of dimensionality} in nearest neighbor problems~\cite{indyk1998approximate}.

Besides~\cite{rahimi2007random}, there have been several approaches proposed to approximate other kernels such as group invariant~\cite{li2010random}, intersection~\cite{maji2009max}, and RBF kernels~\cite{sreekanth2010generalized}. More recently, there has been an interest in approximating polynomial kernels using random feature maps~\cite{kar2012random} and random tensor products~\cite{pham2013fast}. Our work builds on these approaches and provides a more compact representation of approximating polynomial kernels more accurately. 
\section{Compact Random Feature Maps}


We begin by demonstrating that previous approaches for approximating polynomial kernels~\cite{kar2012random}~\cite{pham2013fast} construct rank-deficient spaces. As a solution to this challenges, we present the framework of CRAFTMaps, followed by proving their error bounds and explaining how to generate them efficiently using randomized Hadamard transform.

\subsection{Preliminaries}
\begin{figure}[b]
\centering
    \includegraphics[width=0.5\textwidth]{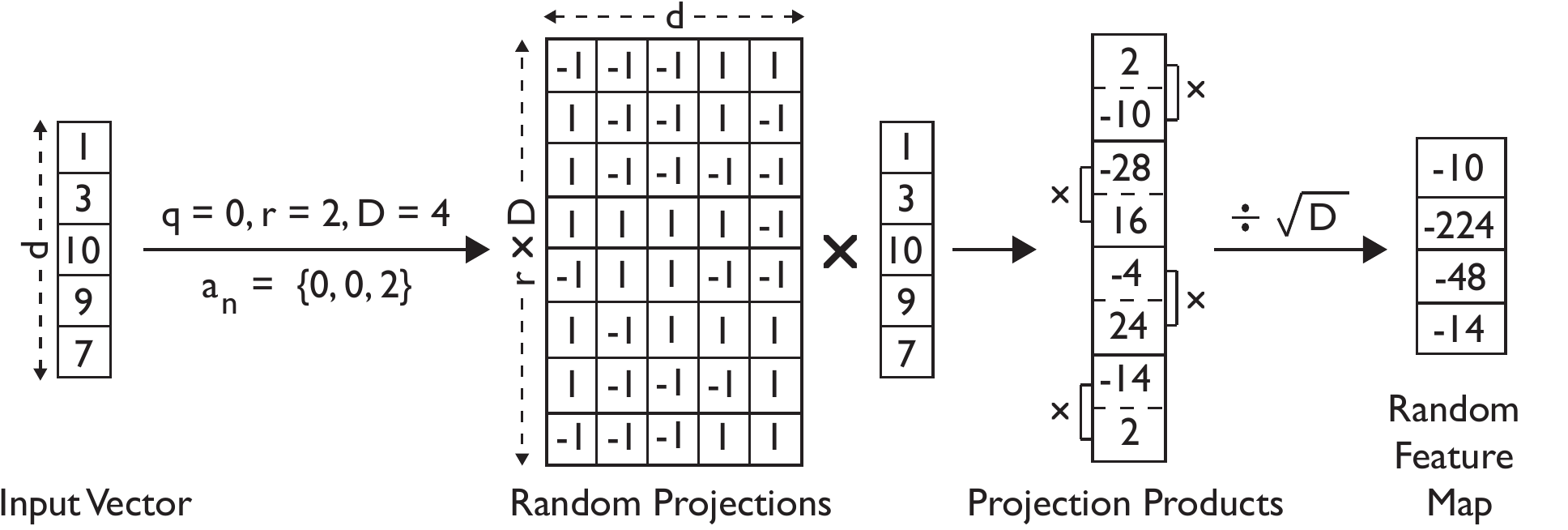}
  \caption{\small{shows Algorithm~\ref{alg:ran_maps} projecting a $5$ dimensional input vector to a random feature map for a $2^{nd}$ order homogenous polynomial kernel in $4$ dimensions.}}
\label{fig:kar_illust}
\end{figure}

 Following~\cite{kar2012random}, consider a positive definite kernel $\textrm{K}:(\mathbf{x},\mathbf{y}) \mapsto f(\langle \mathbf{x},\mathbf{y} \rangle)$, where $f$ admits a Maclaurin expansion with only non-negative coefficients, \textit{i.e.}, $f(x)=\sum_{n=0}^{\infty}a_{n}x^{n}$, where $a_{n} \geq 0$. An example of such a kernel is the polynomial kernel $\textrm{K}(\mathbf{x},\mathbf{y}) = (\langle \mathbf{x}, \mathbf{y} \rangle$$+q)^r$, with $q \in \mathbb{N}_{0}$ and $r \in \mathbb{R}^{+}$. By defining estimators for each individual term of the kernel expansion, one can approximate the exact kernel dot-products. To this end, let $\mathbf{w} \in \mathbb{R}^{d}$ be a Rademacher vector, \textit{i.e.}, each of its components are chosen independently using a fair coin toss from the set $\{ -1, 1\}$. It can be shown that the feature map $Z: \mathbb{R}^d \rightarrow \mathbb{R}^\textrm{D}$, $Z: \mathbf{x} \mapsto \sqrt{a_{\textrm{N}}p^{\textrm{N}+1}}\prod_{j=1}^{\textrm{N}}\mathbf{w}_{j}^{\text{T}}\mathbf{x}$ gives an unbiased estimate of the polynomial kernel. Here $\mathbb{P}[\textrm{N}=n] = 1/(p^{n+1})$, and $\mathbf{w}$$_{1}, \cdot\cdot\cdot, \mathbf{w}_{\textrm{N}}$ are $\textrm{N}$ independent Rademacher vectors. Generating $\textrm{D}$ such feature maps independently and concatenating them together constructs a multi-dimensional feature map $\textbf{Z}: \mathbb{R}^d \rightarrow \mathbb{R}^\textrm{D}, \textbf{Z}: \mathbf{x} \mapsto 1/\sqrt{\textrm{D}} (Z_{1}(\mathbf{x}), \cdot \cdot \cdot, Z_{\textrm{D}}(\mathbf{x}))$, such that $\E{\langle \textbf{Z}(\mathbf{x}), \textbf{Z}(\mathbf{y}) \rangle}=\textrm{K}(\mathbf{x},\mathbf{y})$. The procedure for generating random feature maps for polynomial kernels is listed in Algorithm~\ref{alg:ran_maps} and illustrated in Figure~\ref{fig:kar_illust}.

\begin{algorithm}[t]
\caption{{\sc -- Random Feature Maps (RFM)}} \label{alg:ran_maps}
{\bf Input:} Kernel parameters $q$ and $r$, output dimensionality $\textrm{D}$, sampling parameter $p>0$\\
{\bf Output:} Random feature map $\textbf{Z}:\mathbb{R}^d\rightarrow\mathbb{R}^\textrm{D}$ such that $\langle\textbf{Z}(\mathbf{x}), \textbf{Z}(\mathbf{y})\rangle \approx \textrm{K}(\mathbf{x},\mathbf{y})$\\\vspace{-0.45cm}
\begin{algorithmic}[1]
    \State Set $f(x)=\sum\limits_{n=0}^{\infty}a_{n}x^{n}$ where $a_{n}=\frac{f^{n}(0)}{n!}$
        \ForEach{$i$ = $1$ to $\textrm{D}$}
            \State Set $\textrm{N}\in\mathbb{N}_{0}$ for $P[\textrm{N}=n]=\frac{1}{p^{n+1}}$
            \State Sample $\mathbf{w}_{1},\cdot\cdot\cdot,\mathbf{w}_{\textrm{N}}\in\{-1,1\}^d$
            \State Set $Z_{i}:\mathbf{x}\mapsto \sqrt{a_{\textrm{N}}p^{\textrm{N}+1}}\prod\limits_{j=1}^{\textrm{N}}\mathbf{w}_{j}^{\text{T}}\mathbf{x}$
        \EndFor
        \State Construct $\textbf{Z}:\mathbf{x}\mapsto\frac{1}{\sqrt{\textrm{D}}}(Z_{1},\cdot\cdot\cdot, Z_{\textrm{D}})$
\end{algorithmic}
\end{algorithm}


\subsection{Limitations of Random Feature Maps}
\label{ss:ran_feat_map_limitations}

Random feature maps are an efficient means to approximate the underlying eigen structure of the exact kernel space. However, their efficiency can come at the cost of their rank deficiency. Consider for instance Figure~\ref{fig:rank_def}(a) where the black graph shows the log-scree plot of the exact $7^{\textrm{th}}$ order polynomial kernel ($q=1$) obtained using $1000$ randomly selected set of points from MNIST data. The red graph shows the log-scree plot for the random feature map~\cite{kar2012random} in a $2^{12}$ dimensional space. It can be observed that the red plot is substantially lower than the black one for majority of the spectrum range. Note that this rank deficiency is also true for the space generated by random tensor products~\cite{pham2013fast} whose log-scree plot is shown in green in Figure~\ref{fig:rank_def}(a).

This rank deficiency can result in the under-utilization of the projected feature space. Figure~\ref{fig:rank_def}(b) shows the histogram of the linear weight vector learned in a $2^{12}$ dimensional random feature map~\cite{kar2012random} for a $7^{\textrm{th}}$ order polynomial kernel ($q=1$). The plot was obtained for $1000$ randomly selected points from MNIST data for two class-sets. The spike at zero shows that a majority of the learned weight components do not play any role in classification.

\begin{algorithm}[t]
\caption{{\sc -- CRAFTMaps using RFM}} \label{alg:craftmaps}
        {\bf Input:} Kernel parameters $q$ and $r$, up and down projection dimensionalities $\textrm{D}$ and $\textrm{E}$ such that $\textrm{E} < \textrm{D}$, sampling parameter $p>0$\\
        {\bf Output:} CRAFTMap $\textbf{G}: \mathbb{R}^d \rightarrow \mathbb{R}^{\textrm{E}}$, such that $\langle\textbf{G}(\mathbf{x}), \textbf{G}(\mathbf{y})\rangle \approx \textrm{K}(\mathbf{x},\mathbf{y})$\\\vspace{-0.25cm}
        \begin{algorithmic}[1]
            \State \textbf{Up Project}: Using Algorithm $1$, construct random feature map $\textbf{Z}: \mathbb{R}^d \rightarrow \mathbb{R}^\textrm{D}$, such that $\langle\textbf{Z}(\mathbf{x}), \textbf{Z}(\mathbf{y})\rangle \approx \textrm{K}(\mathbf{x},\mathbf{y})$
            \State \textbf{Down Project}: Using Johnson-Lindenstrauss random projection, linearly down-project $\textbf{Z}$ to construct $\textbf{G}: \mathbb{R}^\textrm{D} \rightarrow \mathbb{R}^\textrm{E}$ such that $\langle \textbf{G}(\textbf{Z}(\mathbf{x})), \textbf{G}(\textbf{Z}(\mathbf{y}) \rangle \approx \langle\textbf{Z}(\mathbf{x}), \textbf{Z}(\mathbf{y})\rangle$.\vspace{0.15cm}
    \end{algorithmic}
\end{algorithm}

\subsection{CRAFTMaps using Up/Down Projections}
\label{ss:up_down_proj}

To address the limitations of random feature maps, we propose CRAFTMaps as a more accurate approximation of polynomial kernels. The intuition behind CRAFTMaps is to first capture the eigen structure of the exact kernel space comprehensively, followed by representing it in a more concise form. CRAFTMaps are therefore generated in the following two steps:

\begin{figure}[h]
\centering
    \includegraphics[width=0.49\textwidth]{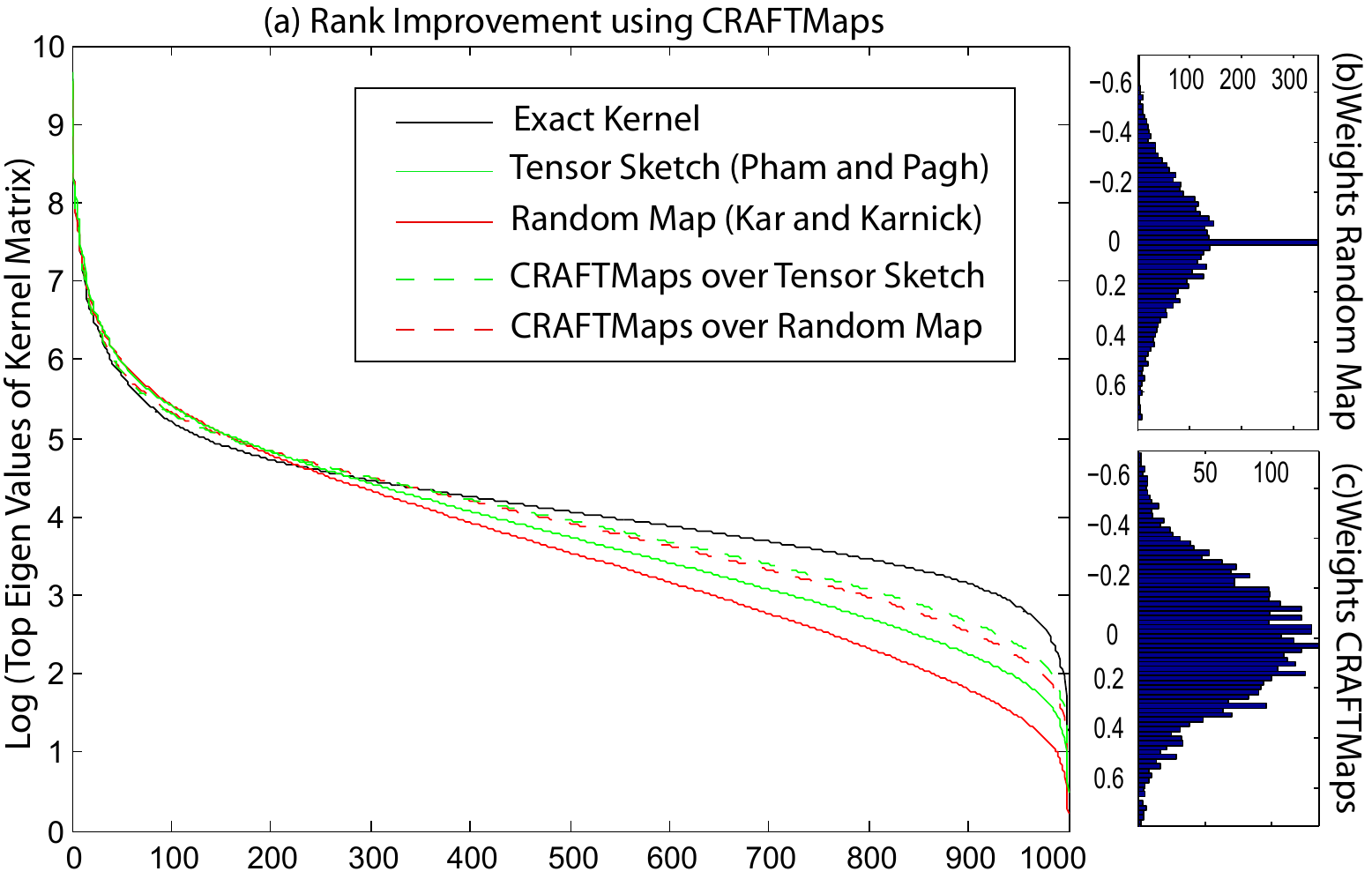}
\caption{\small{\textbf{(a)} Rank deficiency of tensor sketch~\cite{pham2013fast} and random feature maps~\cite{kar2012random}, along with rank improvements due to CRAFTMaps. \textbf{(b-c)} Histograms of weight vectors learned in a $2^{12}$ dimensional random feature map~\cite{kar2012random} and CRAFTMaps (here $\textrm{D}$ was set equal to $2^{14}$).}}\vspace{-0.3cm}
\label{fig:rank_def}
\end{figure}

\textbf{Up Projection:} Since the difference between $\langle \textbf{Z}(\mathbf{x}), \textbf{Z}(\mathbf{y}) \rangle$ and $\textrm{K}(\mathbf{x},\mathbf{y})$ reduces exponentially as a function of the dimensionality of \textbf{Z}~\cite{kar2012random}~\cite{pham2013fast}, we first up project the original data non-linearly from  $\mathbb{R}^{d}$ to a substantially higher dimensional space $\mathbb{R}^{\textrm{D}}$ to maximally capture the underlying eigen structure of the exact kernel space.

\textbf{Down Projection:} Since the randomized feature map $\textbf{Z}: \mathbb{R}^d \rightarrow \mathbb{R}^\textrm{D}$ generated as a result of the up-projection step is fundamentally rank-deficient ($\S$~\ref{ss:ran_feat_map_limitations}), we linearly down project \textbf{Z} to a lower-dimensional map $\textbf{G}: \mathbb{R}^\textrm{D} \rightarrow \mathbb{R}^{\textrm{E}}$, such that $\textrm{E} < \textrm{D}$, and $\langle \textbf{G}(\textbf{Z}(\mathbf{x}))$, $\textbf{G}(\textbf{Z}(\mathbf{y})) \rangle \approx \langle \textbf{Z}(\mathbf{x}), \textbf{Z}(\mathbf{y}) \rangle$. The procedure to generate CRAFTMaps is listed in Algorithm~\ref{alg:craftmaps}. Note that while Algorithm~\ref{alg:craftmaps} uses random feature maps~\cite{kar2012random} for up-projection, one could also use tensor products~\cite{pham2013fast} instead to generate $\textbf{Z}$.

The rank improvement brought about by using CRAFTMaps for random feature maps and tensor sketch is shown in Figure~\ref{fig:rank_def}-a by the dotted red and green plots respectively. The improved utilization of the projected space of random feature maps due to CRAFTMaps is demonstrated in Figure~\ref{fig:rank_def}(c).

\subsection{Error Bounds for CRAFTMaps}
\label{ss:error_bounds}
Recall that the following result obtained using an application of the Hoeffding inequality~\cite{hoeffding1963probability} is central to the analysis of~\cite{kar2012random}:
\begin{equation}
\small
  \prob{ \abs{\ip{\textbf{Z}(\mathbf{x})}{\textbf{Z}(\mathbf{y})} - \textrm{K}(\mathbf{x},\mathbf{y})} > \varepsilon} \le 2 \exp \left( -
  \frac{\textrm{D}\varepsilon^2}{8 \textrm{C}_{\Omega}^2} \right) 
\label{eqn:basic}
\end{equation}
\noindent We first examine this inequality more closely for homogenous polynomial kernels $\textrm{K}(\mathbf{x},\mathbf{y}) = \ip{\mathbf{x}}{\mathbf{y}}^r$ for all points on the unit sphere. In that case we have,
\begin{equation}
  \textrm{C}_\Omega^2 = (p f(p\textrm{R}^2))^{2} = \left(\frac{1}{2^{r+1}}\right)^{2} d^{2r}
\end{equation}
\noindent where $\textrm{R} = \max \norm{\mathbf{x}}_{\ell_1} = \sqrt{d}$ and a suitable choice for $p$ is $1/2$. We only get a non-trivial bound when $\textrm{D} \gtrsim \varepsilon^{-2}d^{2r}$. Note however that if we used explicit kernel expansion, we would need substantially fewer features (at most $\binom{d+r-1}{r}$).  The same holds for~\cite{pham2013fast} since
they apply the same Hoeffding inequality, and the analysis produces the same asymptotics.

We therefore first present an improved error analysis of~\cite{kar2012random}, focusing on homogeneous polynomial kernels. We then use this analysis to prove error bounds of CRAFTMaps. Note that these bounds are independent of the dimensionality of the input space, which is a significant improvement over both~\cite{kar2012random} and~\cite{pham2013fast}.

\begin{lemma}
\label{lm:lemma_1}
\normalfont Fix an integer $r \geq 2$, and define $\textrm{S}_\textrm{D}$ as:
\[
\textrm{S}_\textrm{D} = \sum_{i=1}^\textrm{D} \prod_{j=1}^r \langle \mathbf{x}, \omega_{i,j}\rangle \langle \mathbf{x}^\prime, \omega_{i,j} \rangle
\]
where $\mathbf{x}, \mathbf{x}^\prime$ are vectors of unit Euclidean length, and $\omega_{i,j} \sim \mathcal{N}(0,I_{d})$ are independent Gaussian vectors. Then whenever $\textrm{D} \geq 3 \cdot 4^{r+2} \varepsilon^{-2},$
\[
\prob{\left|\frac{1}{\textrm{D}} \textrm{S}_\textrm{D} - \langle \mathbf{x}, \mathbf{x}^\prime \rangle^r\right| \geq \varepsilon} \leq c^r \exp \left( - \frac{1}{2} \left(\frac{\textrm{D} \varepsilon^2}{11} \right)^{\frac{1}{2r+2}} \right)
\]
where $ 0 < c < 0.766$ is a universal constant.
\end{lemma}
\noindent \textbf{Proof:} Let $\textrm{Y}_i = \prod_{j=1}^r \langle \mathbf{x}, \omega_{i,j} \rangle \langle \mathbf{x}^\prime, \omega_{i,j} \rangle,$ then the deviation of $\textrm{S}_\textrm{D}$ from its mean is estimated by the rate at which the tails of $\textrm{Y}_i$ decay, which is in turn determined by the rates at which the moments of $\textrm{Y}_i$ grow. We first verify that the expectation of the summands indeed equals $\langle \mathbf{x}, \mathbf{x}^\prime \rangle^{r}$:
\[ \E {\textrm{Y}_i} = \prod_{j=1}^r \E {\mathbf{x}\transp \omega_{i,j} \omega_{i,j}\transp \mathbf{x}^\prime} = \langle \mathbf{x}, \mathbf{x}^\prime \rangle^r \]
Similarly, the $k^{\textrm{th}}$ moment of $\textrm{Y}_i$ can be determined as:
  \begin{align*}
  \E {|\textrm{Y}_i|^k} & =
       \prod_{j=1}^r \E{|\tr{\mathbf{x} \transp \omega_{i,j} \omega_{i,j}\transp \mathbf{x}^\prime}|^k} \\
       & \le \prod_{j=1}^r \left[\TNorm{\mathbf{x}^\prime \mathbf{x}\transp}^k \E{\tr{\mathbf{x}\transp \omega_{i,j} \omega_{i,j}\transp  \mathbf{x}}^k }\right]\\
       & = \prod_{j=1}^r \E{|\omega_{i,j}\transp \mathbf{x}|^{2k}} = \prod_{j=1}^r \E{|\gamma_{j}|^{2k}} \\
       & = \left[ \left(\frac{1}{2}\right)^k \frac{(2k)!}{k!}  \right]^r \leq (\sqrt{2})^r \left( \frac{2k}{e} \right)^{rk} \leq c^r k^{rk}
       \end{align*}
 Here $\gamma_j \sim \mathcal{N}(0,1)$, $c = \sqrt{2} (2/e)^k$, and the last three expressions above follow from the formula for the moments of a standard Gaussian random variables~\cite{patel1996handbook}. We now estimate moments of feature map approximation error.
  \[
  \textrm{Q} = \frac{1}{\textrm{D}^k} \E{\Big| \sum_{i=1}^\textrm{D} (\textrm{Y}_i - \E{\textrm{Y}_i}) \Big|^k}
  \]
  Assuming $k \geq 2$, and using Marcinkiewicz--Zygmund inequality~\cite{Burkholder88} we have:
  \[
  \textrm{Q} \leq \left(\frac{k}{\sqrt{\textrm{D}}}\right)^k \E{|\textrm{Y}_i - \E{\textrm{Y}_i}|^k}
  \]
  A standard estimate of the right-hand quantity using Jenson's inequality allows us to conclude that
  \[
  \textrm{Q} \leq \left(\frac{2k}{\sqrt{\textrm{D}}}\right)^k \E{|Y_i|^k} \leq c^r \left(\frac{2k}{\sqrt{\textrm{D}}}\right)^k k^{rk}
  \]

  Finally, we apply Markov's inequality to bound the tails of the approximation error:
  \begin{align*}
  \prob{\left|\frac{1}{\textrm{D}} \sum_{i=1}^\textrm{D} \textrm{Y}_i - \langle \mathbf{x}, \mathbf{x}^\prime \rangle^r \right| \geq \varepsilon} & \leq \frac{\textrm{Q}}{\varepsilon^k} \leq c^r \left(\frac{2k}{\varepsilon \sqrt{\textrm{D}}}\right)^{k}k^{rk} \\
   & \hspace{-6.0em} = c^r \exp\left( k \big[ \log(2k^{r+1}) - \log(\varepsilon \sqrt{\textrm{D}}) \big] \right)
  \end{align*}

   Fixing $\alpha > 0$ and assuming that $\textrm{D} > e^{2 \alpha} 4^{r+2} \varepsilon^{-2}$ and $k = \lfloor (\varepsilon^2 \textrm{D} e^{-2\alpha}/4)^{1/(2r+2)} \rfloor$ ensures that
  \[
  \log(2k^{r+1}) - \log(\varepsilon \sqrt{\textrm{D}}) \leq -\alpha
  \]
  and $k \geq 2,$ so our earlier assumption when applying Marcinkiewicz--Zygmund inequality is valid. Thus
  \[
 \prob{\left|\frac{1}{\textrm{D}} \sum_{i=1}^\textrm{D} \textrm{Y}_i - \langle \mathbf{x}, \mathbf{x}^\prime \rangle^r\right| \geq \varepsilon} \leq c^r \exp \left(- \alpha \left(\frac{\textrm{D} \varepsilon^2}{4 e^{2\alpha}} \right)^{\rho}\right)
 \]
where $\rho = 1/(2r+2)$ and $c \leq \sqrt{2} (2/e)^2 < 0.766.$ Take $\alpha = 1/2$ to reach the bound in the theorem.$\hfill \square$

\noindent Applying Lemma~\ref{lm:lemma_1}, the following corollary follows:

\begin{corollary}
\normalfont  Let $\textrm{X} \subset \R^d$ be a set of $n$ unit vectors. Let $\mathbf{\omega}_{i,j}
  \sim N(0,I_d)$ be a set of $r \cdot \textrm{D}$ independent Gaussian random
  vectors. If $\textrm{D} \gtrsim 4^{r+1} \log(n)^{2r+2}\varepsilon^{-2}$ then we have with high
  probability:
\begin{align*}
    \abs{\frac{1}{\textrm{D}} \sum_{i=1}^\textrm{D} \prod_{j=1}^r \ip{\mathbf{x}}{\mathbf{\omega}_{i,j}}
      \ip{\mathbf{x}'}{\mathbf{\omega}_{i,j}} - \ip{\mathbf{x}}{\mathbf{x}'}^r} \le \varepsilon
  \end{align*}
\end{corollary}

\noindent which holds simultaneously $\forall$ $\mathbf{x}, \mathbf{x}' \in \textrm{X}$.

\noindent \textbf{Proof:} We apply the Lemma~\ref{lm:lemma_1} along with the trivial union bound over
  $\mathcal{O}(n^2)$ points. Thus, we require $\exp( \log(n^2) - (\textrm{D}\varepsilon^2)^{1/(2r+2)})$ to be small. In this case, picking $\textrm{D} \ge \textrm{log}(n^{2})^{(2r+2)}\varepsilon^{-2}$ suffices.$\hfill \square$

An alternate way to view this is to fix $\textrm{D}$, in which case the final approximation error will be bounded by:
\begin{equation}
\varepsilon \lesssim \textrm{log}(n^{2})^{r+1}/\sqrt{\textrm{D}}
\label{eq:d_bound}
\end{equation}
\noindent We can combine this with a usual Johnson-Lindenstrauss~\cite{johnson1984extensions} random projection as follows:

\begin{theorem}
\normalfont  Let $\textrm{X} \subset \R^d$ be a set of $n$ unit vectors. Suppose we map
  these vectors using a random feature map $\mathbf{Z}: \R^d \to \R^\textrm{D}$
  composed with a Johnson-Lindenstrauss map $\textrm{Q}: \R^\textrm{D} \to \R^\textrm{E}$,
  where $\textrm{D} \ge \textrm{E}$, to obtain $\mathbf{Z}'$, then the following holds:
  \begin{align*}
    \abs{ \ip{\textbf{x}}{\textbf{x}'}^r - \ip{\mathbf{Z}'(\mathbf{x})}{\mathbf{Z}'(\mathbf{y})}} \lesssim \frac{2^{r+1}
      \log(n)^{r+1}}{\textrm{D}^{1/2}} + \frac{\log(n)^{1/2}}{\textrm{E}^{1/2}}
  \end{align*}
 with high probability $\forall$ $\mathbf{x}, \mathbf{x}' \in \textrm{X}$ simultaneously.
\end{theorem}

\noindent \textbf{Proof:} A Johnson-Lindenstrauss projection from $\mathbb{R}^{\textrm{D}}$ to $\mathbb{R}^{\textrm{E}}$ preserves with high probability all pairwise inner products of the $n$ points $\{\textbf{Z}(\textbf{x}): \textbf{x} \in \textbf{X}\}$ in $\mathbb{R}^{\textrm{D}}$ to within an additive factor of $\varepsilon' \lesssim \log(n)^{1/2} / \textrm{E}^{1/2}$. Applying the triangle inequality:
\begin{flalign*}
    &\abs{ \ip{\mathbf{x}}{\mathbf{y}}^r - \ip{\mathbf{Z}'(\mathbf{x})}{\mathbf{Z}'(\mathbf{y})}} \le \abs{ \ip{\mathbf{x}}{\mathbf{y}}^r - \ip{\mathbf{Z}(\mathbf{x})}{\mathbf{Z}(\mathbf{y})}} + & \\
    &\abs{ \ip{\mathbf{Z}(\mathbf{x})}{\mathbf{Z}(\mathbf{y})} - \ip{\mathbf{Z}'(\mathbf{x})}{\mathbf{Z}'(\mathbf{y})}} := \varepsilon + \varepsilon' &
\end{flalign*}

Referring to Equation~\ref{eq:d_bound} to bound $\varepsilon$, we obtain the final error bound:
\begin{align*}
  \varepsilon + \varepsilon' \lesssim \frac{2^{r+1} \log(n)^{r+1}}{\textrm{D}^{1/2}} +
  \frac{\log(n)^{1/2}}{\textrm{E}^{1/2}}
\end{align*}
$\hfill \square$

\noindent In particular, the error is lower than random feature maps~\cite{kar2012random}
whenever:
\begin{align*}
  \frac{2^{r+1}\log(n)^{r+1}}{\textrm{D}^{1/2}} + \frac{\log(n)^{1/2}}{\textrm{E}^{1/2}} \lesssim \frac{2^{r+1}\log(n)^{r+1}}{\textrm{E}^{1/2}}
\end{align*}
Fixing $\textrm{D} = g(r) \textrm{E}$ for some constant $g(r) \ge 1$, CRAFTMaps provide a better error bound when:
\begin{align*}
  g(r) \gtrsim \left(\frac{\textrm{log}(n)^{r+1/2}}{\textrm{log}(n)^{(r+1/2)} - 2^{-(r+1)}}\right)^{2} \approx 1
\end{align*}

\subsection{Efficient CRAFTMaps Generation}
\label{ss:e_craftmaps}
Recall that for Hessian based optimization of linear regression problems, the dominant cost of $\mathcal{O}(n\textrm{D}^{2}$) is spent calculating the Hessian. By compactly representing random feature maps in $\mathbb{R}^\textrm{E}$ as opposed to $\mathbb{R}^\textrm{D}$ for $\textrm{E} < \textrm{D}$, CRAFTMaps provide a factor of $\textrm{D}^{2}/\textrm{E}^{2}$ gain in the complexity of Hessian computation. A straightforward version of CRAFTMaps would incur an additional cost of $\mathcal{O}(n\textrm{D}\textrm{E}$) for the down-projection step. However, since for problems at scale $n >> \textrm{D}$, the gains CRAFTMaps provide for classifier learning over random feature maps is well worth the relatively small additional cost they incur.

These gains can be further improved by using structured random matrices for the up/down projections of CRAFTMaps. One way to do this is to use the Hadamard matrix as a set of orthonormal bases, as opposed to using a random bases-set sampled from a zero mean distribution. The structured nature of Hadamard matrices enables efficient recursive matrix-matrix multiplication that only requires $\mathcal{O}(n^{2}\textrm{log}(n))$ operations compared to the $\mathcal{O}(n^{3})$ operations needed for the product of two $n \times n$ non-structured matrices. Constructing CRAFTMaps using Hadamard transform can therefore reduce the complexity of up projection from $\mathcal{O}(n\textrm{D}d)$ to $\mathcal{O}(n\textrm{D}\textrm{log}(d))$, and that of down projection from $\mathcal{O}(n\textrm{D}^{2})$ to $\mathcal{O}(n\textrm{D}\textrm{log}(\textrm{D}))$ respectively. To employ Hadmard matrices for efficient CRAFTMaps generation, we use the sub-sampled randomized Hadamard transform (SRHT)~\cite{tropp2011improved}.

\begin{figure}[t]
\centering
    \includegraphics[width=0.5\textwidth]{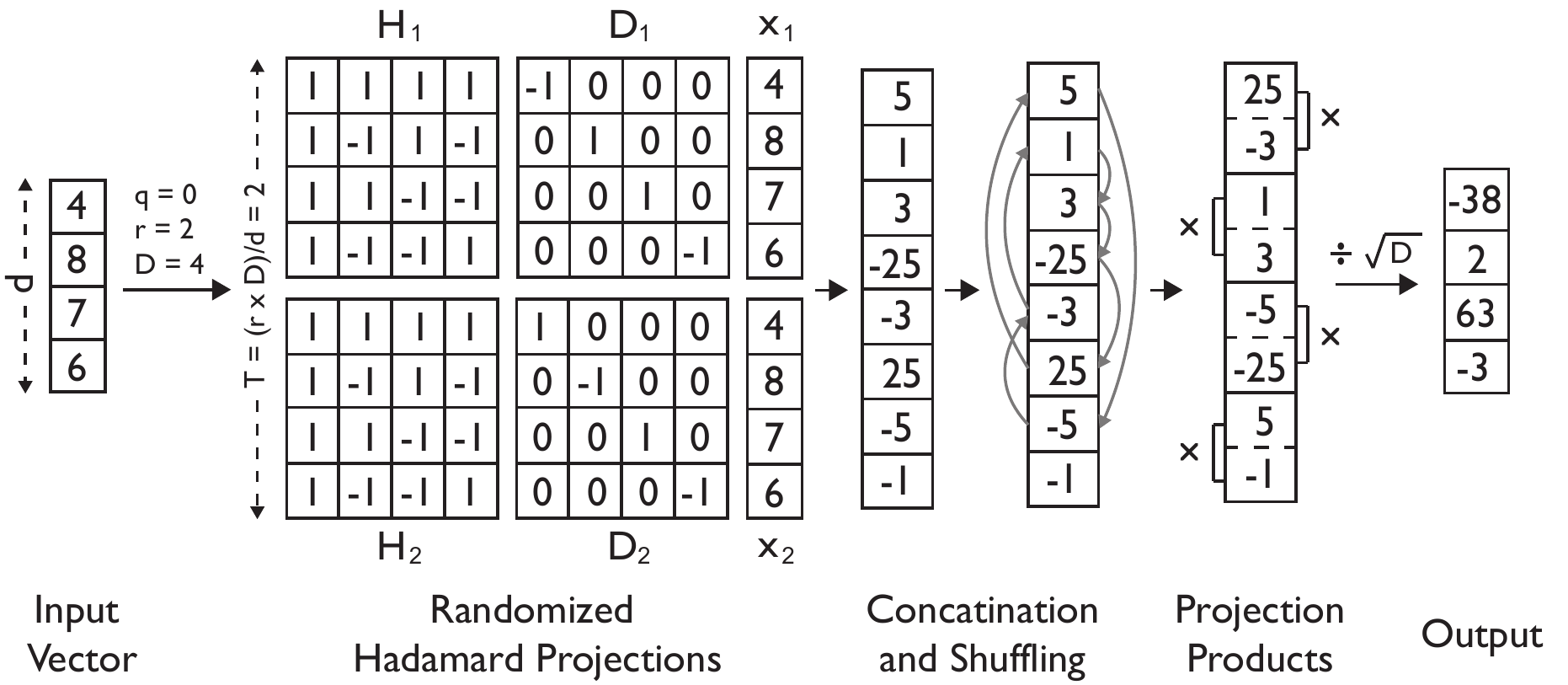}
  \caption{\small{The figure shows the illustration of using randomized Hadamard basis for up-projecting an input vector in $4$ dimensional space to a random map for a $2^{nd}$ order homogenous kernel in a $4$ dimensional space.}\vspace{-0.5cm}}
\label{fig:had_ecocs}
\end{figure}

While SRHT can be used directly for the down-projection step, we need to incorporate a few novel modifications to it before it can be used for up-projection. In particular, given a kernel function $\textrm{K}:(\mathbf{x},\mathbf{y}) \mapsto f(\langle \mathbf{x},\mathbf{y} \rangle)$ and a $d$ dimensional\footnote{As Hadamards exist in powers of $2$, usually $\mathbf{x}$ needs to be zero-padded to the closest higher power of $2$.} vector $\mathbf{x}$, we first construct $\textrm{T} = \lceil\sum_{i=1}^D \textrm{N}_{i})/d\rceil$ copies of $\mathbf{x}$, where $\textrm{N}$ is defined in Algorithm~\ref{alg:ran_maps}. Each copy $\mathbf{x}_{t}$ is multiplied by a diagonal matrix $\mathbf{M}_{t}$ whose entries are set to $+1$ or $-1$ with equal probability. Each matrix $\mathbf{M}_{t}\mathbf{x}_{t}$ is implicitly multiplied by the $d \times d$ Hadamard matrix $\mathbf{H}$. All rows of $\mathbf{H}\mathbf{M}_{t}\mathbf{x}_{t}$ for all $t = \{1, \cdot\cdot\cdot, \textrm{T}\}$ are first concatenated, and then randomly permuted, to be finally used according to Algorithm~\ref{alg:ran_maps} to non-linearly up-project $\mathbf{x}$ from $\mathbb{R}^{d}$ to $\mathbb{R}^{\textrm{D}}$ (see Figure~\ref{fig:had_ecocs}).

\section{Classification Using ECOCs}

To solve multi-class classification problems, we use error correcting output codes (ECOCs)~\cite{dietterich1995solving} which employ a unique binary ``codeword" of length $c$ for each of the $k$ classes, and learn $c$ binary functions, one for each bit position in the codewords. For training, using an example from class $i$, the required outputs of the $c$ binary functions are specified by the codeword for class $i$. Given a test instance $\mathbf{x}$, each of the $c$ binary functions are evaluated to compute a $c$-bit string $s$. This string is compared to the $k$ codewords, assigning $\mathbf{x}$ to the class whose codeword is closest to $s$ according to some distance. 

Overall, given $d$ dimensional data from $k$ classes, we first use up/down projections to construct its CRAFTMap representation in $\mathbb{R}^\textrm{E}$. We then use the framework of ECOCs to learn $c$ binary linear regressors in $\mathbb{R}^\textrm{E}$. We perform multi-fold cross validation on the training data to select one regularization parameter $\lambda$ that is used for all the $c$ codeword classifiers. To test a $d$ dimensional example, it is first up/down projected to $\mathbb{R}^\textrm{E}$, and then passed through ECOCs to be classified to one of the $k$ classes.

\section{Experiments and Results}
\label{sec:experiments}

\begin{figure*}[t]
\centering
\includegraphics[width=1.0\textwidth]{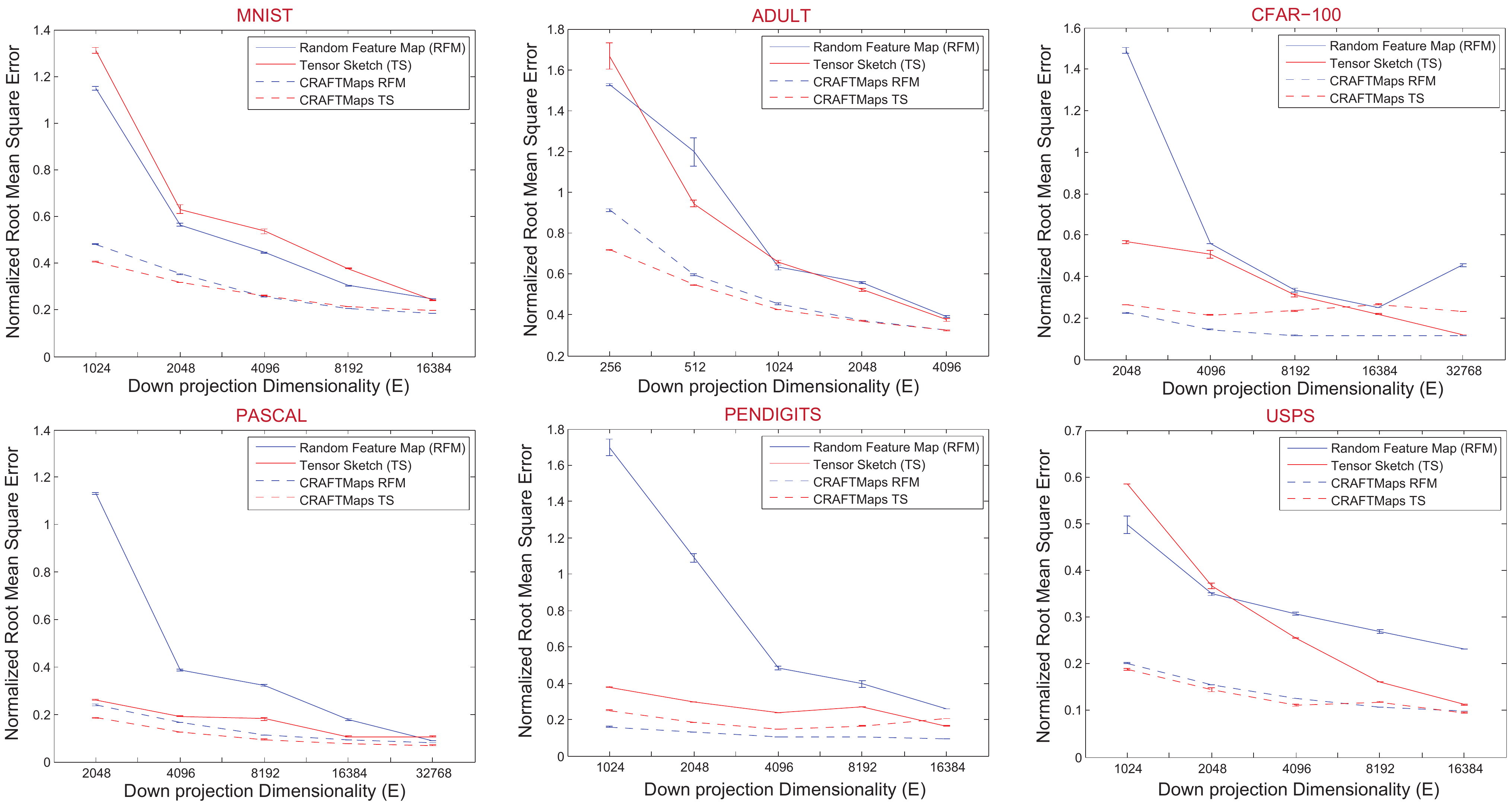}
  \caption{Normalized root mean square (nrms) errors obtained while reconstructing the polynomial kernel with $r=7$ and $q=1$ using random feature maps~\cite{kar2012random}  and tensor sketching~\cite{pham2013fast} versus CRAFTMaps. Results over $6$ data-sets are presented. For each plot, $\textrm{D}$ was set at twice of $\textrm{max(E)}$.}
  \label{fig:reconst_results}
\end{figure*}

We now present  reconstruction and classification results of CRAFTMaps on multiple data-sets. 

\subsection{Reconstruction Error}
\label{ss:reconst_error}

Figure~\ref{fig:reconst_results} shows the normalized root mean square (nrms) errors obtained while reconstructing the polynomial kernel with $r=7$ and $q=1$ using random feature maps~\cite{kar2012random}  and tensor sketching~\cite{pham2013fast} versus their respective CRAFTMap representations. Results over $6$ different data-sets are presented. All graphs in each plot were obtained using $10$ folds of $1000$ randomly selected data points from a particular data-set. As shown, CRAFTMaps provide a significant reconstruction improvements for random maps and tensor sketching.

\begin{figure}[t]
\centering
\includegraphics[width=0.345\textwidth]{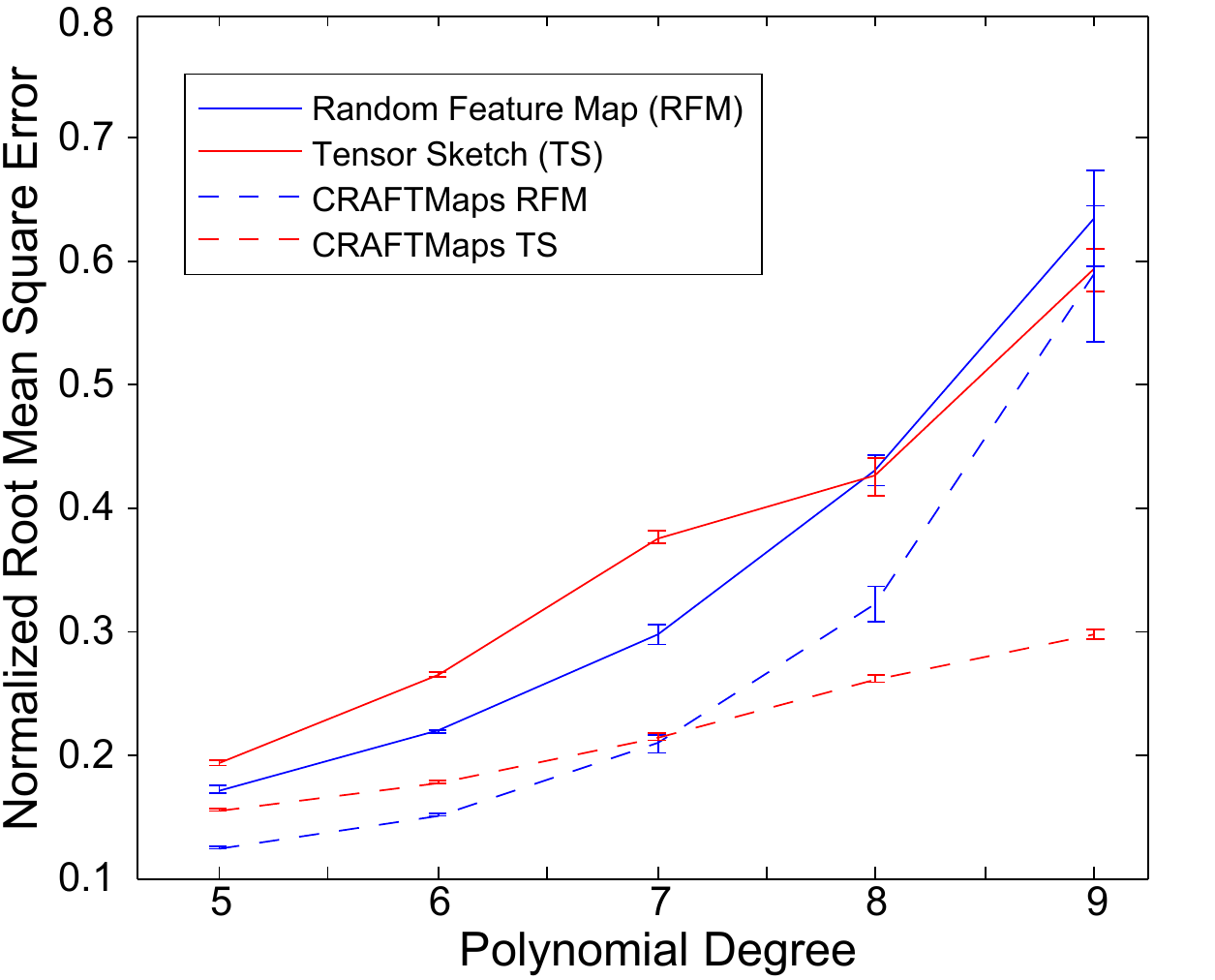}
  \caption{Reconstruction error as a function of polynomial degree, averaged over $10$ randomly sampled $1000$ points of MNIST data. Here $\mathbb{R}^{\textrm{D}} = 2^{15}$ while $\mathbb{R}^{\textrm{E}} = 2^{13}$.}\vspace{-0.4cm}
  \label{fig:reconst_results_a}
\end{figure}

Figure~\ref{fig:reconst_results_a} shows the reconstruction improvements due to CRAFTMaps as a function of polynomial degree. These results were obtained using $10$ sets of $1000$ randomly picked points from MNIST data. As shown, CRAFTMaps consistently improve the reconstruction error over a range of polynomial degrees.

\subsection{Classification Error}
\label{ss:class_error}

Table~\ref{tab:mnist_results} shows the test classification errors obtained using random feature maps~\cite{kar2012random} and tensor sketching~\cite{pham2013fast} versus their CRAFTMap representations. Results over $4$ different data-sets are presented, on which CRAFTMaps consistently delivered improved classification performance.

We now explain results for CRAFTMaps on MNIST data for small and substantially large projected feature spaces. We also explain CRAFTMaps results on very large amounts of training data using MNIST8M.

\textbf{Small Feature Spaces:} Table~\ref{tab:mnist_results}-\textbf{a} shows MNIST results on feature space sizes $300$ to $700$ dimensions. Note that for $\mathbb{R}^{\textrm{E}} < d$ (which for MNIST is $784$ ), the random feature maps cannot use the $\textbf{H}$-$0/1$ heuristic of~\cite{kar2012random}. CRAFTMaps however do not have this limitation as even for $\textrm{E} < d$, $\textrm{D}$ can still be $>>d$. This allows CRAFTMaps to use the $\textbf{H}$-$0/1$ heuristic in $\mathbb{R}^{\textrm{D}}$, which in turn reflects in $\mathbb{R}^\textrm{E}$. This results in substantial classification gains achieved by CRAFTMaps for small-sized feature spaces, and highlights their usefulness in applications with low memory footprint such as mobile phone apps.

\textbf{Large Feature Spaces:} Table~\ref{tab:mnist_results}-\textbf{b} shows the MNIST results on feature space sizes $2^{12}$ to $2^{16}$ dimensions. It can be seen that CRAFTMaps consistently gave improved test error and achieved $\textbf{1.12\%}$ test classification rate using the original $60$K training data (unit-length normalized, non-jittered and non-deskewed).

\begin{table}[h]
\begin{centering}
  \begin{tabular}{c}
    \hspace{-0.15cm}\subfloat{\taba}\\
    \hspace{-0.1855cm}\subfloat{\tabb}\\
    \hspace{-0.2125cm}\subfloat{\tabc}\\
    \hspace{-0.2125cm}\subfloat{\tabd}\\
    \hspace{-0.35cm}\subfloat{\tabe}\\
  \end{tabular}
  \end{centering}
  \caption{Test classification errors for $4$ data-sets for $r =$ $7$, $5$, $5$ and $9$ respectively and $q = 1$. Here RFM and TS stand for~\cite{kar2012random} and~\cite{pham2013fast}. The first row of each table shows values of $\textrm{E}$, while $\textrm{D}$ was set equal to $8$ times $\textrm{E}$.}\vspace{-0.3cm}
  \label{tab:mnist_results}
\end{table}

\begin{figure}[h]
\centering
\includegraphics[width=0.4\textwidth]{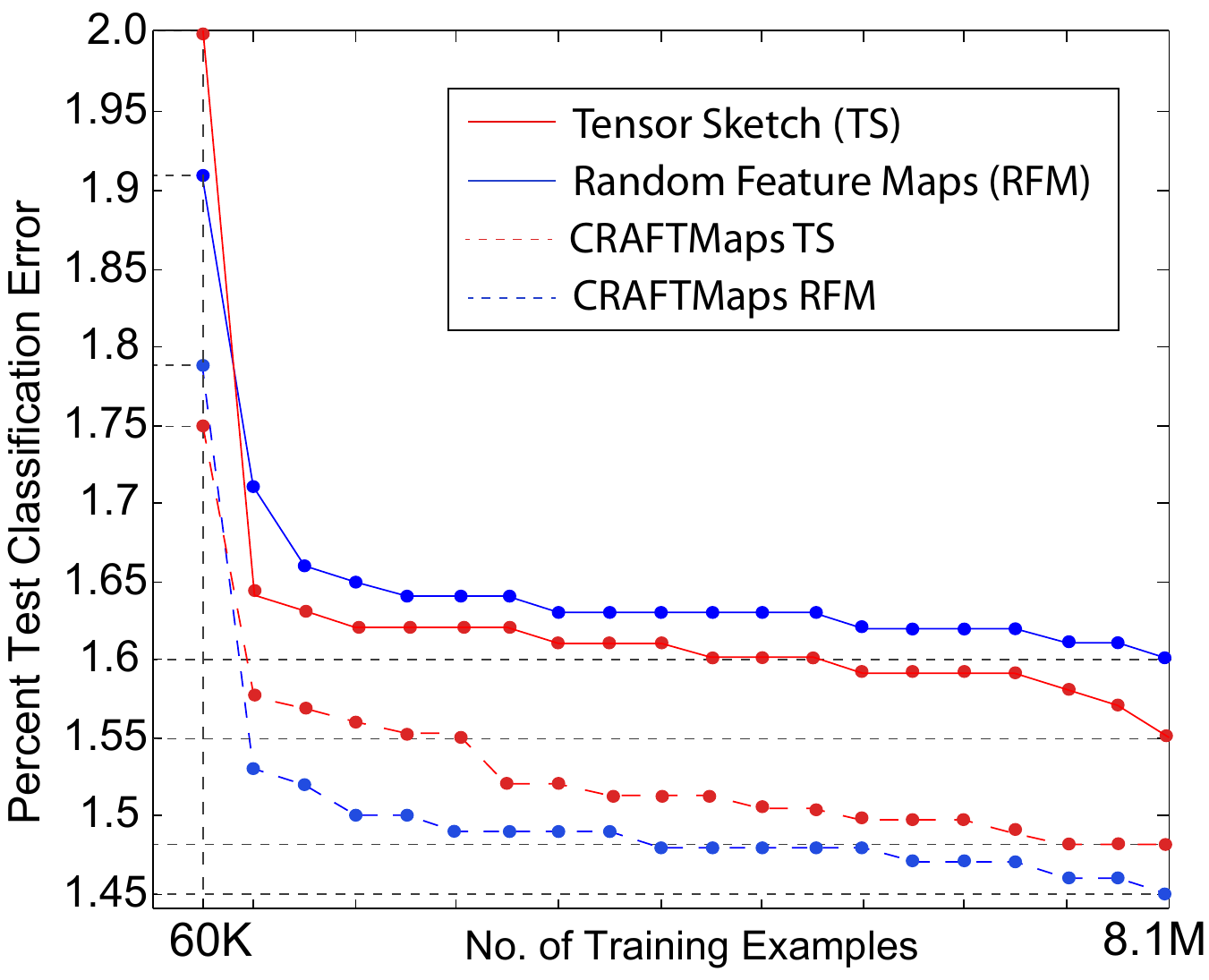}
  \caption{Test classification for MNIST8M. Here $\textrm{D} = 2^{17}$, $\textrm{E} = 2^{14}$, $q = 1$ and $r = 7$ and ECOCs = $200$.}\vspace{-0.15cm}
  \label{fig:mnist_8m}
\end{figure}

\begin{figure}[h]
\centering
\includegraphics[width=0.415\textwidth]{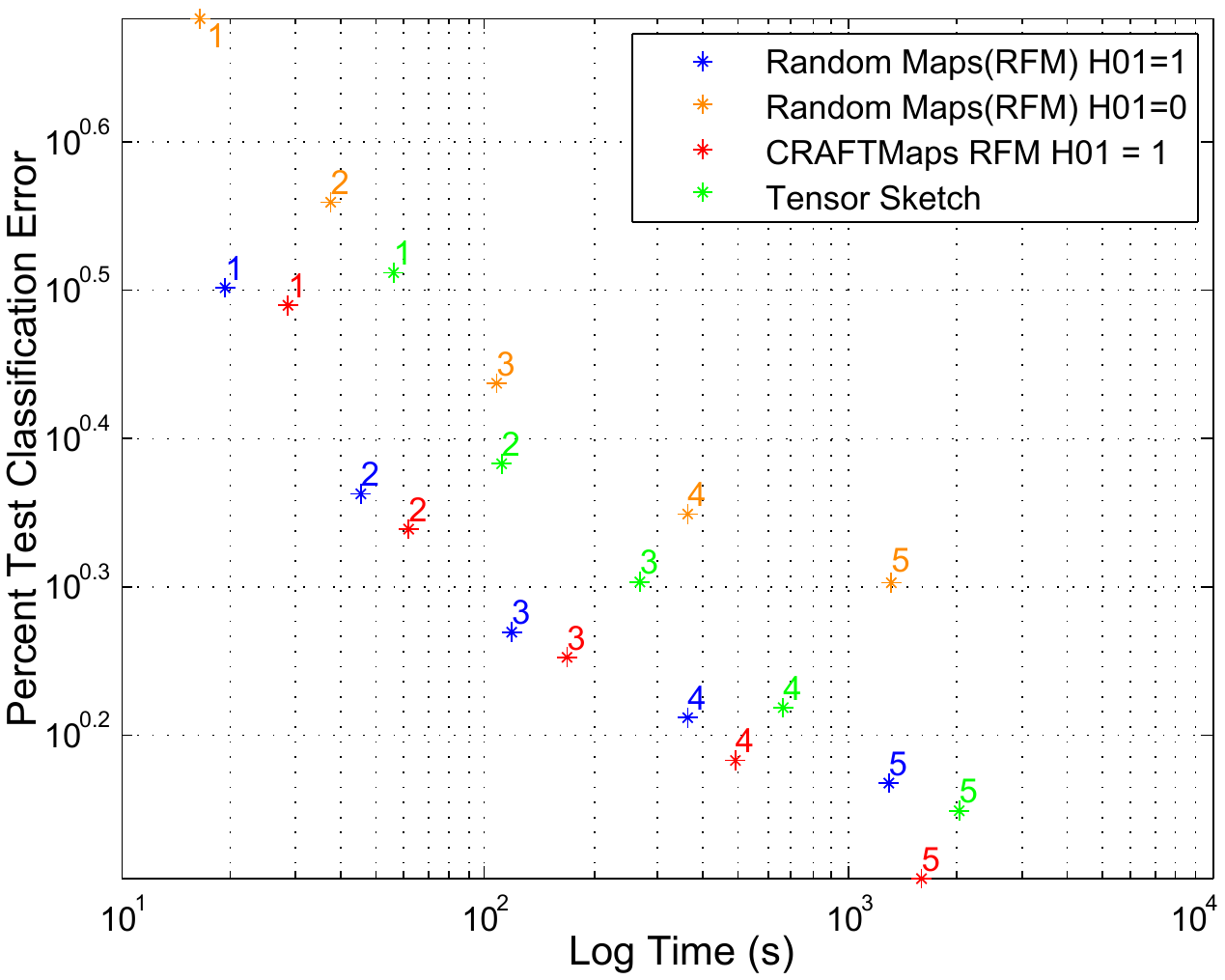}
  \caption{Log-log scatter plot of compute times (projection
+ Hessian) for MNIST data.}\vspace{-0.2cm}
  \label{fig:time_analysis}
\end{figure}

\textbf{Results on MNIST$8$M Data} Figure~\ref{fig:mnist_8m} shows  the performance of CRAFTMaps in comparison to random feature maps for a given sized $\mathbb{R}^{\textrm{E}}$ ($2^{14}$) as the number of examples vary from $60$ thousand to $8.1$ million. This experiment uses the same set of $10$ thousand test points as used for the experiments with MNIST data. It can be seen that CRAFTMaps on random feature maps converge the fastest, and consistently gives better classification performance compared to the other representations. These results were obtained using a polynomial kernel with $r=7$, $q=1$, $\textrm{D}=2^{17}$, $\textrm{E}=2^{14}$, and ECOCs equal to $200$. As we increase $\textrm{E}$ to $2^{16}$ and $\textrm{D}$ to $2^{19}$ using CRAFTMaps on RFM for $7^{\textrm{th}}$ order polynomial kernel ($q=1$), we achieved test classification error of $\textbf{0.91\%}$ on MNIST$8$M data-set.


\subsection{Run-Time Analysis}
\label{ss:time_analysis}

Figure~\ref{fig:time_analysis} shows the log-log scatter plot of the compute times (projection + Hessian) for random feature maps~\cite{kar2012random}, tensor sketching~\cite{pham2013fast}, and CRAFTMaps using random feature maps (with H-01 heuristic). These times were recorded for MNIST data using a $40$-core machine. Notice that CRAFTMaps show significant per unit-time classification improvements towards the right end of the x-axis. This is because as the size of the projected space increases, the Hessian computation cost becomes dominant. This naturally gives CRAFTMaps an edge given their ability to encode information more compactly. The performance gain of CRAFTMaps are expected to grow even more as training size increases. 
\section{Conclusions and Future Work}

In this work, we proposed CRAFTMaps to approximate polynomial kernels more concisely and accurately compared to previous approaches. We theoretically proved error bounds of CRAFTMaps and presented empirical results to demonstrate their effectiveness.

An important context where CRAFTMaps are particularly useful is the map-reduce setting. By computing a single Hessian matrix (with different gradients for each ECOC) in a concise feature space, CRAFTMaps provide an effective way to learn multi-class classifiers in a single-pass over large amounts of data. Moreover, their ability to compactly capture the eigen structure of the kernel space makes CRAFTMaps suitable for smaller scale applications such as mobile phone apps.



\clearpage

\bibliography{icml_2014}
\bibliographystyle{icml2014}

\end{document}